\documentclass[letterpaper]{article}
\usepackage{proceed2e}
\usepackage[margin=1in]{geometry}
\usepackage{apacite}
\usepackage{times}
\usepackage{color}

\title{Understanding Measures of Uncertainty for Adversarial Example Detection}
\usepackage{placeins}
\author{} 

\author{ {\bf Lewis Smith
} \\
Department of Engineering Science \\
University of Oxford\\
Oxford, United Kingdom\\
\And
{\bf Yarin Gal}  \\
Department of Computer Science \\
University of Oxford\\
Oxford, United Kingdom\\
}

\usepackage{microtype}
\usepackage{graphicx}
\usepackage{natbib}
\usepackage{subfigure}
\usepackage{booktabs} 

\usepackage{amsmath}
\usepackage{amssymb}
\allowdisplaybreaks
\usepackage[small,compact]{titlesec}
\titlespacing{\section}{0pt}{1ex}{0.8ex}
\titlespacing{\subsection}{0pt}{0.5ex}{0ex}
\titlespacing{\subsubsection}{0pt}{0.2ex}{0ex}
\expandafter\def\expandafter\normalsize\expandafter{%
 \normalsize
 \setlength\abovedisplayskip{2pt}
 \setlength\belowdisplayskip{2pt}
 \setlength\abovedisplayshortskip{0pt}
 \setlength\belowdisplayshortskip{0pt}
}

\begin{document}

\maketitle

\begin{abstract}Measuring uncertainty is a promising technique for
  detecting adversarial examples, crafted inputs on which the model
  predicts an incorrect class with high confidence. But many measures of uncertainty
  exist, including predictive entropy and mutual information, each
  capturing different types of uncertainty. We study these
  measures, and shed light on why mutual information seems to be
  effective at the task of adversarial example detection. We 
  highlight failure modes for MC dropout, a widely used approach for
  estimating uncertainty in deep models. This leads
  to an improved understanding of the drawbacks of current methods,
  and a proposal to improve the quality of uncertainty estimates using
  probabilistic model ensembles. We give illustrative experiments
  using MNIST to demonstrate the intuition underlying the different
  measures of uncertainty, as well as experiments on a real-world Kaggle dogs
  vs cats classification dataset.

\end{abstract}

\section{Introduction}
\label{introduction}

Deep neural networks are state of the art models for representing complex, high
dimensional data such as natural images. However, neural networks are
not robust: it is possible to find small
perturbations to the input of the network that produce erroneous and
over-confident classification results. Such perturbed inputs, known as
adversarial examples \citep{szegedy2013intriguing}, are a major hurdle
for the use of deep networks in safety-critical applications, or those
for which security is a concern.

\begin{figure}[b!]
  \vspace{-6mm} \centering \includegraphics[width=\columnwidth,
  trim=0mm 6mm 0mm 16mm, clip]{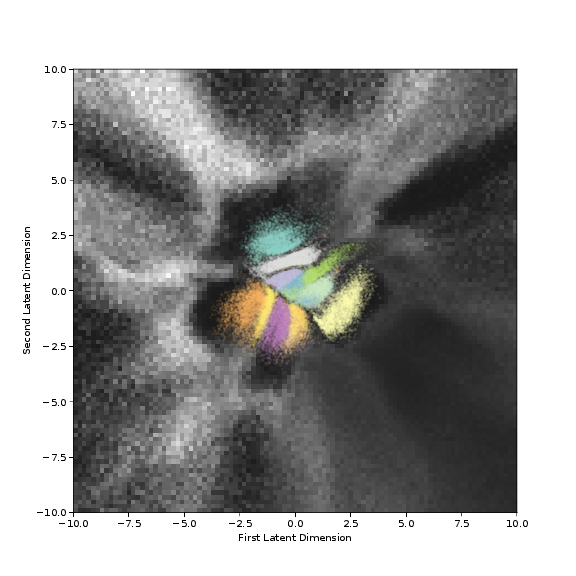}
  \vspace{-8mm}
  \caption{
    Uncertainty of a standard dropout network trained on MNIST, as measured by 
    \textit{mutual information}, visualized in the latent space obtained from a variational
    autoencoder. Colours are classes for each encoded training image.
    In white is uncertainty, calculated by decoding each latent point
    into image space, and evaluating the mutual information between
    the decoded image and the model parameters. A lighter background
    corresponds to higher uncertainty.}
  \label{fig:ml_latent_space_uncertainty}
  \vspace{-2mm}
\end{figure}

One possible hypothesis for the existence of adversarial examples is
that such images lie off the manifold of natural images, occupying
regions where the model makes unconstrained extrapolations.
If this hypothesis were to hold true, then
one would be able to detect adversarial perturbation by measuring the
distance of the perturbed input to the image manifold.

Hypothetically, such distances could be measured using nearest
neighbour approaches, or by assessing the probability of the input
under a density model on image space. However,
approaches based on geometric distance are a
poor choice for images, as pixel-wise distance is a poor metric for
perceptual similarity; similarly, density modelling is difficult to
scale to the very high dimensional spaces found in image recognition.

Instead, we may consider proxies to the distance from the image
manifold. For example, the model uncertainty of a
\textit{discriminative} Bayesian classification model will be high for points far
away from the training data, in theory. High capacity Bayesian neural
networks, for example, could then be used to obtain such measures of
uncertainty. Under the hypothesis that adversarial examples lie far
from the image manifold, i.e.\ the training data, such uncertainty
could be used to identify an input as adversarial.

The uncertainty of such models is not straightforward to obtain. 
Numerical methods for integrating the posterior, such as Markov Chain
Monte Carlo, are difficult to scale to large datasets
\citep{gal2016uncertainty}. As a result, approximations have been studied extensively. For example,
approximate inference in Bayesian neural networks using dropout is a
computationally tractable technique \citep{gal2016dropout} which has
been successfully used in the literature \citep{Leibig2017,
  gal2016uncertainty}. Further, it has been shown that dropout based
model uncertainty can form a good proxy for the detection of
adversarial examples \citep{li2017dropout, feinman2017detecting, rawat2017adversarial}.

However, existing research has mostly overlooked the effect of the chosen \textit{measure
for uncertainty quantification}. Many such measures exist, including
mutual information, predictive entropy, softmax variance, expected
entropy, and the entropy of deterministic deep
networks. \citep{li2017dropout} for example use expected entropy, \citep{rawat2017adversarial} use mutual information,
whereas \citep{feinman2017detecting} estimate the variance of multiple
draws from the predictive distribution (obtained using
dropout). Further, to date, research for the identification of
adversarial examples using model uncertainty has concentrated on toy
problems such as MNIST, and has not been
shown to extend to more realistic data distributions and larger models
such as ResNet \citep{he2015deep}.

In this paper we examine the differences between the various
measures of uncertainty used for adversarial example detection, and in the process provide further evidence for the hypothesis that model uncertainty could be used to identify an input as adversarial. More specifically, we illustrate the differences between the measures by projecting the
uncertainty onto lower dimensional spaces (see for example Fig.\
\ref{fig:ml_latent_space_uncertainty}). 
We show that the softmax variance can be seen as an approximation to the mutual information (section \ref{uncertaintymeasures}), explaining the effectiveness of this rather ad-hoc technique.
We show that non-adversarial off-manifold images (for example image interpolations) are treated in the same way as adversarial inputs by some measures of uncertainty (inputs which were not assessed in previous literature using such measures).
We highlight ways in which dropout
fails to capture the full Bayesian uncertainty by visualizing gaps in model uncertainty in the
latent space (Section \ref{visualisation}), and use this insight to
propose a simple extension to dropout schemes to be studied in future research. We finish by demonstrating the effectiveness of dropout  on the real-world ASIRRA \citep{elson2007asirra} cats and dogs classification dataset (Section \ref{ROC}).
Code for the experiments described in this paper is available online
\footnote{https://github.com/lsgos/uncertainty-adversarial-paper}.

\section{Background}
We begin with a brief review of required literature on Bayesian neural
networks and adversarial examples.

\subsection{Bayesian Deep Learning}
\label{bayesianDL}

A deep neural network (with a given architecture) defines a function
\(f : \mathcal{X} \mapsto \mathcal{Y}\) parametrised by a set of
weights and biases \(\omega = \{\mathbf{W}_l, \mathbf{b}_l\}_{l=1}^L\).  These
parameters are generally chosen to minimize some energy function
\(E : \mathcal{Y} \times \mathcal{Y} \mapsto \mathbb{R} \) on the
model outputs and the target outputs over some dataset
$\mathcal{D} = \{\mathbf{x}_i, \mathbf{y}_i\}_{i=1}^N$ with
$\mathbf{x} \in \mathcal{X}$ and $\mathbf{y} \in \mathcal{Y}$. Since
neural networks are highly flexible models with many degrees of
freedom, modifications to this loss to regularize the solution
are often necessary, so that the weights are chosen as
\begin{align}
  \hat{\omega} = \arg \min_{\omega} \sum_i E(f(\mathbf{x}_i; \omega), y) + \lambda \sum_l || \mathbf{W}_l ||^2
  \label{map}
\end{align}
for the common choice of an \(L_2\) regulariser with weight decay
\(\lambda\).

Rather than thinking of the weights as fixed
parameters to be optimized over, the Bayesian approach is to treat them as random variables,
and so we place a prior distribution \(p(\omega)\) over the weights of
the network.  If we also have a likelihood function
\(p(y \mid \mathbf{x}, \omega)\) that gives the probability of
\(\mathbf{y} \in \mathcal{Y}\) given a set of parameter values and an
input to the network, we can then conduct inference given a dataset by
marginalizing the parameters. Such models are known as \textit{Bayesian neural networks}.

If, as is often the case, the prior on the weights is a zero mean
Gaussian with diagonal covariance, and the energy of the network is
the negative log likelihood (so
\(p(\mathbf{y} \mid \omega, \mathbf{x}) = e^{- E(f(\mathbf{x}), \mathbf{y}) }\)) then
the optimized solution in equation \ref{map} corresponds to a mode
of the posterior over the weights.

Ideally we would integrate out our uncertainty by taking the expectation
of the predictions over the posterior, rather than using this point
estimate.  For neural networks this can
only be done approximately. Here we discuss one practical
approximation, variational inference with dropout approximating
distributions.

\subsection{Variational Inference and Dropout}
\label{varinf}
\textit{Variational inference} is a general technique for
approximating complex probability distributions. The idea is to
approximate the intractable posterior \(p(\omega \mid \mathcal{D})\)
with a simpler approximating distribution \(q_{\theta}(\omega)\).  By
applying Jensen's inequality to the Kullback-Leibler divergence
between the approximating distribution and the true posterior, we can
obtain the log-evidence lower bound \(\mathcal{L}_{VI}\)
\begin{align*}
  \mathcal{L}_{VI} :=& \int q_{\theta} (\omega) \log p (\mathcal{D} \mid \omega) d\omega - D_{KL}(q_{\theta} \mid\mid p(\omega)) .
\end{align*}

Since the model evidence is a constant that does not depend on
the parameters of \(q_{\theta}\), maximizing \(\mathcal{L}_{VI}\) with
respect to \(\theta\) will minimize the KL divergence between \(q\)
and the model posterior. The key advantage of this from a
computational point of view is that we have replaced an integration
problem with the optimization problem of maximising a parametrised
function, which can be approached by standard gradient based techniques.

For neural networks, a common approximating distribution
is \textit{dropout} \citep{srivastava2014dropout} and
it's variants. In the variational framework, this means the weights
are drawn from
\begin{align*}
  \mathbf{W}_l &= \mathbf{M}_l \cdot \text{diag}( [\mathbf{z}_{l,j}]_{j=1}^{K_l}) \\
               &\text{where} \; \mathbf{z}_{l,j} \sim \text{Bernoulli}(p_l), \; l = 1..L, j = 1..K_{l-1} \nonumber
\end{align*}
for a network with \(L\) layers, where the dimension of each layer is
\(K_i \times K_{i-1}\), and the parameters of \(q\) are
\(\theta = \{ \mathbf{M}_l, p_l \mid l = [1..L]\}\). Informally, this
corresponds to randomly setting the outputs of units in the network to
zero (or zeroing the rows of the fixed matrix \(\mathbf{M}_l\)). Often the layer dropout probabilities \(p_i\) are chosen as
constant and not varied as part of the variational framework, but it
is possible to learn these parameters as well \citep{gal2017concrete}.
Using variational inference, the expectation over the posterior can be
evaluated by replacing the true posterior with the approximating
distribution. The dropout distribution is still challenging to
marginalize, but it is readily sampled from, so expectations can be
approximated using the Monte Carlo estimator
\begin{align}
  \mathbb{E}_{p(\omega | \mathcal{D})}[f^\omega(x)] &= \int p(\omega | \mathcal{D}) f^\omega(x) \text{d} \omega \nonumber \\
                          &\simeq \int q_{\theta}(\omega) f^\omega(x) \text{d} \omega \nonumber \\
                          &\simeq \frac{1}{T} \sum_{i=1}^{T} f^{\omega_i}(x), \; \omega_{1..T} \sim q_{\theta}(\omega).
                            \label{dropsample}
\end{align}
\subsection{Adversarial Examples}
\label{adversarial}

Works by \citep{szegedy2013intriguing} and others, demonstrating that state-of-the-art deep image classifiers can be fooled by small perturbations to input images, have initiated a great deal of interest in both understanding the reasons for why such adversarial examples occur with deep models, as well as devising methods to resist and detect adversarial attacks. So far,
attacking has proven more successful than defence; a recent survey of
defences by \citep{carlini2017adversarial} found that, with the partial
exception of the method based on dropout uncertainty analysed by
\citep{feinman2017detecting}, all other investigated defence methods failed.

There is no precise definition of when an example qualifies as
`adversarial'. The most common definition used is of an input
\(\mathbf{x}_{adv}\) which is close to a real data point
\(\mathbf{x}\) as measured by some \(L_p\) norm, but is classified
wrongly by the network with high score. Speaking more loosely, an adversarially
perturbed input is one which a human observer would assign a certain class, but for which the network would predict a different class with a high score.

It is notable that there exists a second, related, type of
images which have troubling implications for the robustness of deep
models, namely meaningless images which are nevertheless classified
confidently as belonging to a particular class (see, for example,
\citep{nguyen2015deep}). These are often produced more for the
visualization of network features than in the interests of producing
an adversarial attack, yet they are still an interesting shortcoming
of neural networks from the point of view of uncertainty, since they
are points that are far from all training data by any reasonable
metric (based on either pixel-wise or perceptual distance), yet the
model gives confident (low predictive entropy) predictions for. We
refer to these as `rubbish class examples' or `fooling images'
following \citep{nguyen2015deep} and \citep{goodfellow2014explaining}.

Several possible explanations for the existence of adversarial
examples have been proposed in the literature
\citep{akhtar2018threat}. These include the idea, proposed in the
original paper by \citep{szegedy2013intriguing}, that the set of
adversarial examples are a dense, low probability set like the
rational numbers on \(\mathbb{R}\), with the discontinuous boundary
somehow due to the strong non-linearity of neural networks. Contrary
to that, \citep{goodfellow2014explaining} proposed that adversarial
examples are partially due of the intrinsically linear response of
neural network layers to their inputs.  \citep{tanay2016boundary} have
proposed that adversarial examples are possible when the decision
boundaries are strongly tilted with respect to the vector separating
the means of the class
clusters. 

Many of these ideas are consistent with the idea that the training
data of the model lies on a low dimensional manifold in image space,
the hypothesis we build upon and provide evidence for in this paper.

\subsection{Measures of Uncertainty}
There are two major sources of uncertainty
a model may have:
\begin{enumerate}
\item \textit{epistemic} uncertainty is uncertainty due to our lack of
  knowledge; we are uncertain because we lack understanding. In terms
  of machine learning, this corresponds to a situation where our model
  parameters are poorly determined due to a lack of data, so our
  posterior over parameters is broad.
\item \textit{aleatoric} uncertainty is due to genuine stochasticity
  in the data. In this situation, an uncertain prediction is the best possible prediction. This corresponds to
  \textit{noisy} data; no matter how much data the model has seen, if
  there is inherent noise then the best prediction possible may be a
  high entropy one (for example, if we train a model to predict coin
  flips, the best prediction is the max-entropy distribution
  \(P(\text{heads}) = P(\text{tails})\)).
\end{enumerate}
In the classification setting, where the output of a model is a conditional
probability distribution \(P(y | x)\) over some discrete set of outcomes
\(\mathcal{Y}\), one straight-forward measure of uncertainty is
the entropy of the predictive distribution
\begin{align}
  H[P(y | x)] = - \sum_{y \in \mathcal{Y}} P(y | x) \log P(y | x).
\end{align}
However, the predictive entropy is not an entirely satisfactory
measure of uncertainty, since it does not distinguish between epistemic and aleatoric uncertainties. However, it may be of interest to do so; in particular, we want
to capture when an input lies in a region of data space
where the model is poorly constrained, and distinguish this from inputs near the data distribution with noisy labels.

An attractive measure of uncertainty able to distinguish epistemic from aleatoric
examples is the information gain between the model parameters and the
data.  Recall that the mutual information (MI) between two random
variables \(X\) and \(Y\) is given by
\begin{align*}
  I(X,Y) &= H[P(X)] - \mathbb{E}_{P(y)}H [P(X \mid Y)] \\
         &= H[P(Y)] - \mathbb{E}_{P(x)}H [P(Y \mid X)].
\end{align*}
The amount of information we would gain about the model
parameters if we were to receive a  label \(y\) for a new point
\(x\), given the dataset \(\mathcal{D}\) is then given by
\begin{align}
  I(\omega, y \mid \mathcal{D}, x) &= H[p(y\mid x,\mathcal{D})]
                                     - \mathbb{E}_{p(\omega \mid \mathcal{D})} H[p(y \mid x, \omega)]
                                     \label{MI}
\end{align}
Being uncertain about an input point $x$ implies that if we knew the
label at that point we would gain information. Conversely, if the parameters at a point are already well
determined, then we would gain little information from obtaining the label. Thus,
the MI is a measurement of the model's \textit{epistemic} uncertainty.

In the form presented above, it is also readily approximated using the
Bayesian interpretation of dropout.  The first term we will refer to
as the `predictive entropy'; this is just the entropy of the
predictive distribution, which we have already discussed. The second
term is the mean of the entropy of the predictions given the
parameters over the posterior distribution
\(p(\omega \mid \mathcal{D})\), and we thus refer to it as the
expected entropy.

These quantities are not tractable analytically for deep nets, but
using the dropout approximation and equation \eqref{dropsample},
the predictive distribution, entropy and the MI are readily computable
\citep{gal2016uncertainty}:
\begin{align}
  p(y \mid \mathcal{D}, \mathbf{x}) &\simeq \frac{1}{T} \sum_{i=1}^{T} p(y \mid \omega_i, \mathbf{x}) \\
                                    &:= p_{MC}(y \mid  \mathbf{x}) \nonumber \\
  H[ p(y \mid \mathcal{D}, \mathbf{x}) ] &\simeq H[ p_{MC}(y \mid \mathcal{D}, \mathbf{x})] \\
  \label{predent}
  I(\omega, y \mid \mathcal{D}, x) &\simeq H[ p_{MC}(y \mid \mathcal{D}, \mathbf{x})] \\
                                     &\qquad - \frac{1}{T} \sum_{i=1}^{T}  H[p(y \mid \omega_i, \mathbf{x})]
\end{align}
where \(\omega_i \sim q(\omega \mid \mathcal{D})\) are samples from
the dropout distribution.

Other, more ad-hoc, measures of uncertainty include the variance of the softmax probabilities $p(y=c \mid \omega_i, \mathbf{x})$ (with the variance calculated over $i$), and variation ratios \citep{gal2016uncertainty}, with the former commonly used in previous research.

In the next section we shed light on the properties of these measures of uncertainty in the context of adversarial example detection. We relate the more ad-hoc softmax variance measure to the principled mutual information, and give intuition into why some measures are more suitable for adversarial example detection than others. This is followed by empirical evaluation of these ideas.

\section{Understanding Measures of Uncertainty for Adversarial Example Detection}

We start by explaining the type of uncertainty relevant for adversarial example detection under the hypothesis that adversarial images lie off the manifold of natural images, occupying regions where the model makes unconstrained extrapolations. 
We continue by relating the fairly ad-hoc \textit{softmax variance} measure of uncertainty to mutual information. 

\subsection{Adversarial Examples and Uncertainty}

Both the MI and predictive entropy should increase on inputs which lie far from the image manifold. Under our hypothesis, we expect both to be effective in highlighting such inputs. However, predictive entropy could also be high \textit{near} the image manifold, on inputs which have inherent ambiguity. Such inputs could be ambiguous images (for example, with MNIST, an image that could be either of class 1 or 7), or more generally interpolations between classes. Such inputs would have high predictive probability for more than one class, yielding high predictive entropy (but low MI). Such inputs are clearly not adversarial, but will falsely trigger an automatic detection system\footnote{We speculate that previous research using predictive entropy has not encountered this phenomenon due to insufficient coverage of the test cases.}. We demonstrate this experimentally in the next section.

Further, adversarial example crafting algorithms seek to
create an example image with a different class to the original,
typically by either minimising the predicted probability of the current class for 
an untargeted attack, or maximising the predicted probability of a target class.  
This has the side-effect that adversarial examples seek to minimise the
entropy of the predictions, a simple consequence of the normalisation of
the probability.
It is interesting to highlight that this also affects the uncertainty as measured by MI; since both the
mutual information and entropy are strictly positive, the mutual
information is bounded above by the predictive entropy (see equation
\ref{MI}). Therefore, the model giving low entropy predictions at a
point is a sufficient condition for the mutual information to be low
as well. Equally, the mutual information bounds the entropy from
below; it is not possible for a model to give low entropy predictions
when the MI is high.

\subsection{Mutual Information and Softmax Variance}
\label{uncertaintymeasures}

Some works in the literature estimate the epistemic uncertainty of a dropout model using the estimated variance of the MC samples,
rather than the mutual information \citep{Leibig2017,feinman2017detecting,carlini2017adversarial}.
This is somewhat ad-hoc, but seems to work fairly well in practice. 
We suggest a possible explanation to the effectiveness of this ad-hoc measure, arguing that the softmax variance can be seen as a proxy to the mutual information. 

One way to see the relation between the two measures of uncertainty is
to observe that the variance is the leading term in the series expansion of the mutual information which we
demonstrate below.  For brevity, we denote the sampled distributions
\(p(y \mid \omega_i, \mathbf{x})\) as \(p_i\) and the mean predictive
distribution \( p_{MC}(y \mid \mathbf{x})\) as \(\hat{p}\).  These are
in general distribution over \(C\) classes, and we denote the
probability of the \(j^{th}\) class as \(\hat{p}_j\) and \(p_{ij}\)
for the mean and \(i^{th}\) sampled distribution respectively. The
variance score is the mean variance across the classes
\begin{align}
  \hat{\sigma}^2 &= \frac{1}{C} \sum_{j=1}^{C} \frac{1}{T} \sum_{i=1}^{T} (p_{ij} - \hat{p}_j)^2 \\
                 &= \frac{1}{C} \left( \sum_{j=1}^{C} \left( \frac{1}{T} \sum_{i=1}^{T} p_{ij}^2 \right)  - \hat{p}_j^2 \right) \nonumber
\end{align}

And the mutual information score is
\begin{align}
  \hat{I} &= H(\hat{p}) - \frac{1}{T} \sum_i H(p_i) \nonumber \\
          &= \sum_j\left( \frac{1}{T} \sum_i  p_{ij} \log p_{ij} \right)-  \hat{p}_j \log \hat{p}_j \nonumber
\end{align}

Using a Taylor expansion of the logarithm,
\begin{align}
  \hat{I} &= \sum_j \left( \frac{1}{T} \sum_i  p_{ij} (p_{ij} - 1) \right) -  \hat{p}_j (\hat{p}_j - 1)  + ... \nonumber \\
          &= \sum_j \left( \frac{1}{T} \sum_i  p_{ij}^2 \right)-  \hat{p}_j^2  - \left( \frac{1}{T} \sum_i p_{ij} \right) + \hat{p}_j  + ... \nonumber \\
          &= \sum_j^C \left( \frac{1}{T} \sum_i^T  p_{ij}^2 \right)-  \hat{p}_j^2 + ...
\end{align} 
we see that the first term in the series is identical up to a multiplicative constant to the mean variance of the samples.

This relation between the softmax variance and the mutual information measure could explain the effectiveness of the variance in detecting adversarial examples encountered by \citep{feinman2017detecting}. MI increases on images far from the image manifold and not on image interpolations (on which the predictive variance increases as well), with the variance following similar trends.

We next give an empirical study of these various measures of uncertainty, and demonstrate experimentally the claims above.

\section{Empirical Evaluation}
In the next section we demonstrate the effectiveness of various measures of uncertainty as proxies to distance from the image manifold. 
We demonstrate the difference in behaviour between the predictive entropy and mutual information on image interpolations, both for interpolations in the latent space as well as interpolations in image space.
We continue by visualising the various measures of uncertainty, highlighting the differences discussed above. This is further developed by highlighting shortcomings with current approaches for uncertainty estimation, to which we suggest initial ideas on how to overcome and suggest new ideas for attacks (to be explored further in future research). We finish by assessing the ideas discussed in this paper on a real world dataset of cats vs dogs image classification.

\subsection{Measures of Uncertainty on Image Interpolations}

\begin{figure}[t!]
\vspace{-4mm}
  \centering
  \includegraphics[width=\columnwidth]{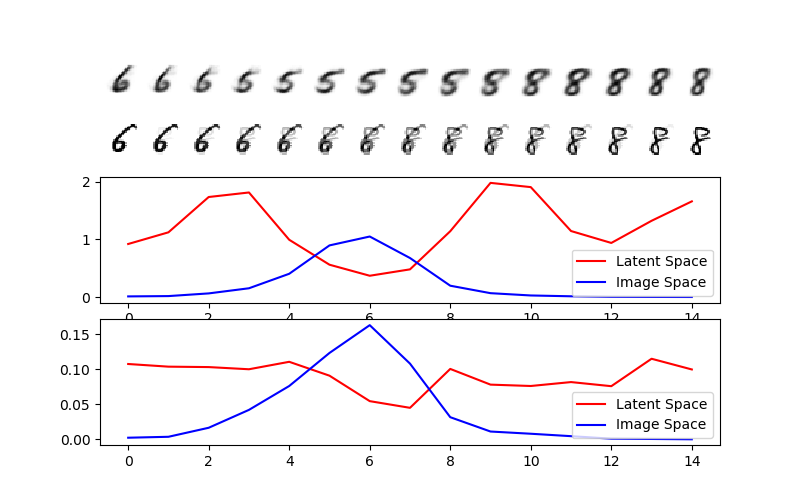}
  \vspace{-8mm}
  \caption{ Entropy (middle) and the MI (bottom) vary along a convex
    interpolation between two images in latent space and image space
    (top).  The entropy is high for regions along
    both interpolations, wherever the class of the image is ambiguous.
    In contrast, the MI is roughly constant along the interpolation in
    latent space, since these images have aleatoric uncertainty (they
    are ambiguous), but the model has seen data that resembles them.
    On the other hand, the MI has a clear peak as the pixel space
    interpolation produces out-of-sample images between the classes
     }
  \label{fig:mi_interpolation}
  \vspace{-2mm}
\end{figure}

\begin{figure}[t!]
\vspace{-4mm}
  \centering \includegraphics[width=0.9\columnwidth,
  trim=0mm 6mm 0mm 16mm, clip]{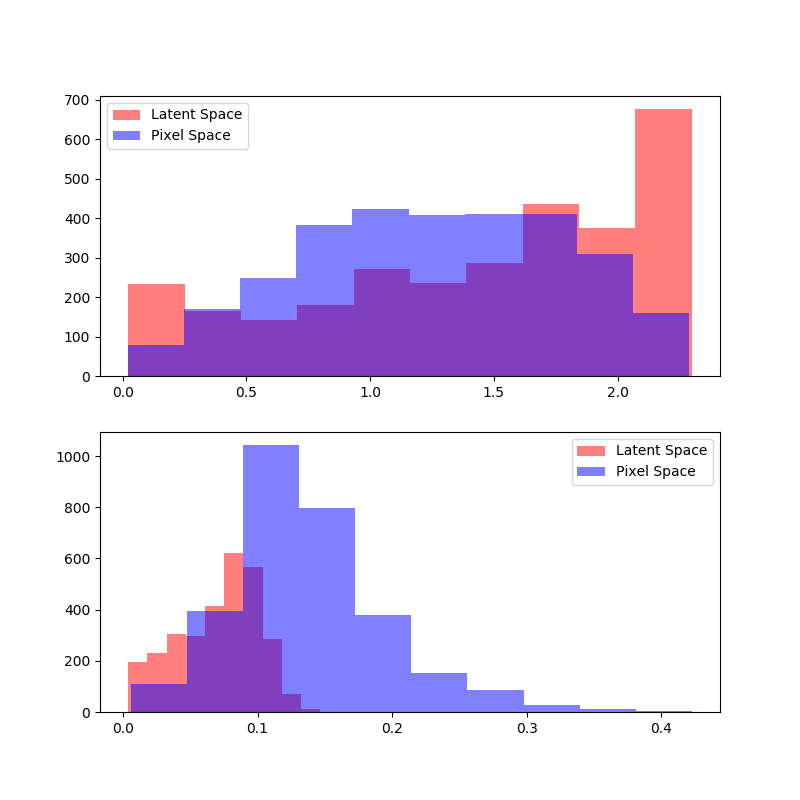}
  \vspace{-8mm}
  \caption{ The entropy (top) and mutual information (bottom) of the
    interpolation halfway between 3000 random points of different
    classes in the MNIST test set, showing the intuition in Figure
    \ref{fig:mi_interpolation} that the two modes of interpolation
    have very different statistical properties with respect to the
    model uncertainty, while the entropy cannot distinguish between
    them.  }
  \vspace{-6mm}
  \label{fig:latent_population}
\end{figure}

We start by assessing the behaviour of the measures of uncertainty on image interpolations, with interpolations done both in latent space, as well as we in image space.
That model uncertainty can capture what we want in practice is
demonstrated in Figures \ref{fig:mi_interpolation} and
\ref{fig:latent_population}, both showing experiments comparing
interpolations, which goes off the manifold of natural images, to
those in the latent space of an auto-encoder, which we assume does a
reasonably good job of capturing the manifold. We see that the MI
distinguishes between these on-manifold and off-manifold images,
whereas the entropy fails to do so. This is necessary for the hypothesis
proposed in the introduction; if we are able to accurately capture the
MI, it would serve well as a proxy for whether an images belongs to the learned
manifold or not.

\subsection{Visualization in Latent Space and Dropout Failures}
\label{visualisation}

\begin{figure}[t!]
  \centering \vspace{-2mm} \includegraphics[width=\columnwidth,
  trim=0mm 6mm 0mm 16mm, clip]{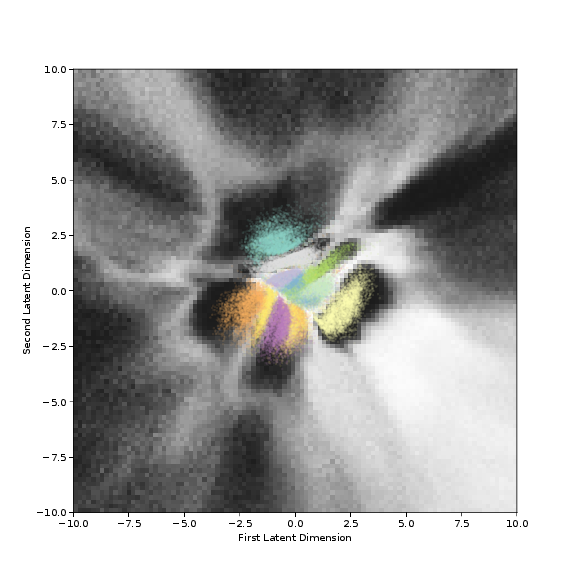}
  \vspace{-8mm}
  \caption{ The predictive entropy of the same network as in figure
  \ref{fig:ml_latent_space_uncertainty}. Note the differences with
    the MI, which is low everywhere close to the data in the centre of
    the plot, but the entropy is high between the classes here.  These
    points correspond to images which resemble digits, but which are
    inherently ambiguous. Note however that there are large regions of
    latent space where the predictive entropy is high and the MI low,
    despite being far from any training data.  }
  \label{fig:ml_latent_space_entropy}
  \vspace{-6mm}
\end{figure}

We wish to gain intuition into how the different measures of uncertainty  behaves. In
order to do so, we use a variational autoencoder \citep{kingma2013auto}
to compress the MNIST 
latent space. By choosing a latent space of two dimensions, we can
plot the encodings of the dataset in two dimensions. By decoding the
image that corresponds to a point in latent space, we can classify the
decoded image and evaluate the network uncertainty, thus providing a
two dimensional map of the input space. Figure \ref{fig:ml_latent_space_uncertainty} shows the latent space with the MI measure of uncertainty, calculated using dropout. Similarly, Figure \ref{fig:ml_latent_space_entropy} shows the predictive entropy measure of uncertainty. Note the differences in uncertainty near the class clusters boundaries (corresponding to image interpolations) -- the MI has low uncertainty in these regions, whereas the predictive entropy is high.

Another question of interest in this context is how well the dropout approximation captures uncertainty.  The
approximating distribution is fairly crude, and variational inference
schemes are known to underestimate the uncertainty of the posterior,
tending to fit an approximation to a local mode rather than capturing
the full posterior\footnote{There are two reasons for this behaviour:
  firstly, that the approximating distribution \(q\) may not have
  sufficient capacity to represent the full posterior, and secondly,
  the asymmetry of the KL divergence, which penalizes \(q\) placing
  probability mass where the support of \(p\) is small far more
  heavily than the reverse.}. 

\begin{figure}[t!]
  \centering \vspace{-2mm} \includegraphics[width=0.7\columnwidth,
  trim=0mm 0mm 0mm 6mm, clip]{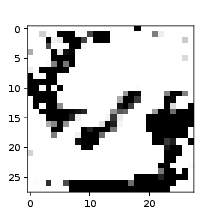}
  \vspace{-6mm}
  \caption{A typical garbage class example generated by searching
    latent space.  This is classified as a 2 with high confidence.}
  \label{fig:latent_garbage}
  \vspace{-6mm}
\end{figure}

As seen from the figures, the network does a reasonable job of
capturing uncertainty close to the data.  However, the network's
uncertainty has `holes'-- regions where the predictions of the model
are very confident, despite the images generated by the decoder here
being essentially nonsense (see Figure \ref{fig:latent_garbage}). This
is somewhat troubling; we want our models to be uncertain on data that
does not resemble what they saw during training. It also suggests
that, while the uncertainty estimates generated by MC dropout are
useful, they do not capture the full posterior, instead capturing local behaviour near one
of it's modes.

\begin{figure}[b!]
  \centering \vspace{-2mm} \includegraphics[width=\columnwidth,
  trim=0mm 6mm 0mm 16mm, clip]{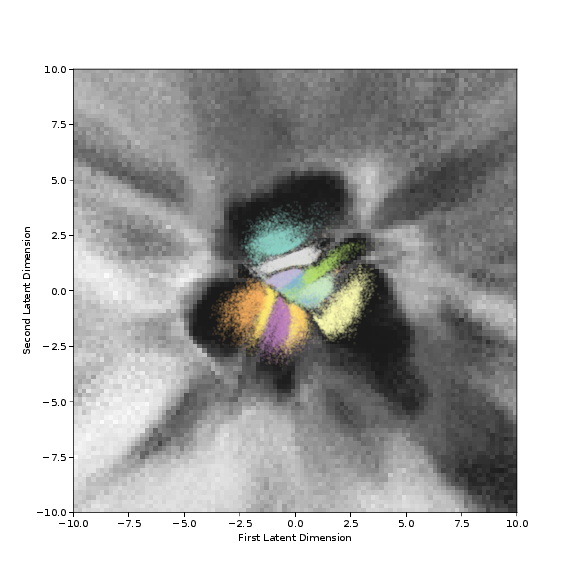}
  \vspace{-8mm}
  \caption{ The MI calculated using an ensemble of dropout models,
    treating all of their predictions as Monte Carlo samples from the
    posterior.  This mitigates some of the spuriously confident
    regions in latent space }
  \label{fig:ensemble_latent_space_entropy}
  \vspace{-6mm}
\end{figure}

This may offer an explanation as to why MC dropout nets are still
vulnerable to adversarial attack; since they do not capture the full
posterior, there are still regions where they are mistakenly
overconfident, which adversarial attack algorithms can exploit. This
intuition suggests a simple fix; since a single dropout model averages
over a single mode of the posterior, we can capture the posterior
using an ensemble of dropout models using different initializations,
assuming that these will converge to different local modes.  We find
that even a small ensemble can qualitatively improve this behaviour
(Figure \ref{fig:ensemble_latent_space_entropy}). We leave further developments of these insights, into both new adversarial example crafting techniques using uncertainty gaps, as well as into new mitigation techniques using ensembles, for future research.

Lastly, it should be noted that there is no guarantee that an ensemble of
dropout models captures the true posterior. It will approximate it
well only if the true posterior is concentrated in many local modes,
all of roughly equal likelihood (since all the models in the ensemble
are weighted equally), and a randomly initialized variational dropout
net trained with some variant of gradient descent will converge to all
of these modes with roughly equal
probability. 

\subsection{Evaluation on Cats and Dogs Dataset}
\label{ROC}

\begin{figure}[b!]
  \centering
  \includegraphics[width=\columnwidth, trim=14mm 14mm 14mm 14mm, clip]{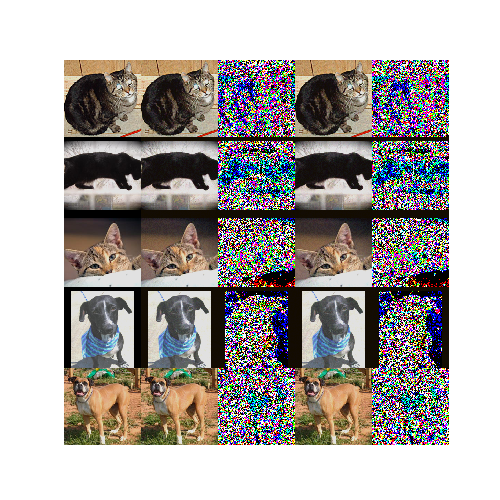}
  \caption{ Example adversarial images generated by the Momentum
    iterative method at \(\epsilon=10\), with original images on the
    left, adversarial images on the deterministic model in the second column,
    and those for the MC dropout model in the fourth column. The difference between
    the adversarial image and the original is shown on the right of each image.}
  \label{fig:adv_examples}
\end{figure}

\begin{table*}[t!]
\vspace{-4mm}
  \centering
  \caption{The AUC for the adversarial discrimination task described in the experiments section. Fields marked with (S) denote this quantity evaluated on a version of the dataset with unsuccessful adversarial examples (that do not change the label) removed. The success rate of each attack in changing the label is given as a measure of each attacks effectiveness on this dataset. }
  \vspace{4mm}
    \begin{sc}
    \begin{tabular}{l|cc|cc|c}
      \toprule
          & \multicolumn{1}{l}{entropy} & \multicolumn{1}{l}{MI} & \multicolumn{1}{l}{entropy (S)} & \multicolumn{1}{l}{MI (S)} & \multicolumn{1}{l}{Success Rate} \\
  \midrule
    bim   \(\epsilon= 5\) &     &       &       &           &  \\
    Deterministic & 0.322 & N.A  & 0.293 & N.A  &  0.757 \\
    MC & 0.0712 & \textbf{0.728} & 0.0617 & \textbf{0.733} &  0.900 \\
    \hline
    fgm   \(\epsilon= 5\)&     &       &       &              &  \\
    Deterministic Model & 0.439& N.A  & \textbf{0.490} & N.A&   0.517 \\
    MC Model & 0.426 & \textbf{0.557} & 0.465 & \textbf{0.497} &  0.563\\
    \hline
    mim   \(\epsilon= 5\)&     &       &       &              &  \\
    Deterministic Model & 0.347 & N.A  & 0.319 & N.A  & 0.743 \\
    MC Model & 0.0476& \textbf{0.657} & 0.0410 & \textbf{0.669} &  0.917 \\
    \hline
    \hline
    bim   \(\epsilon= 10\)&    &       &       &              &  \\
    Deterministic Model & 0.302& N.A  & 0.285& N.A  & 0.753\\
    MC Model & 0.0686 & \textbf{0.708} & 0.0719 & \textbf{0.723} &  0.917 \\
    \hline
    fgm   \(\epsilon= 10\)&    &       &       &             &  \\
    Deterministic Model & 0.502 & N.A  & \textbf{0.550} & N.A  & 0.487 \\
    MC Model & 0.480 & \textbf{0.529} & 0.514 & 0.491 &  0.547 \\
    \hline
    mim   \(\epsilon= 10\)&    &       &       &              &  \\
    Deterministic Model & 0.350 & N.A  & 0.319 & N.A  & 0.763\\
    MC Model & 0.0527& \textbf{0.661}& 0.0442& \textbf{0.665}&  0.907 \\
      \bottomrule
    \end{tabular}%
      \end{sc}
  \label{tab:addlabel}%
\end{table*}%

\begin{figure*}[h]
\vspace{-8mm}
\begin{minipage}[t]{0.32\textwidth}
  \centering
  \includegraphics[width=\columnwidth]{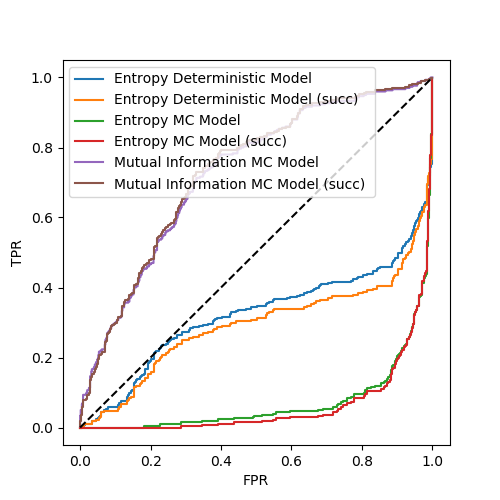}
  \caption{BIM with \(\epsilon=5\)}
  \label{fig:rcc_results_bim_rn50roc_curves}
\end{minipage}
\begin{minipage}[t]{0.32\textwidth}
  \centering
  \includegraphics[width=\columnwidth]{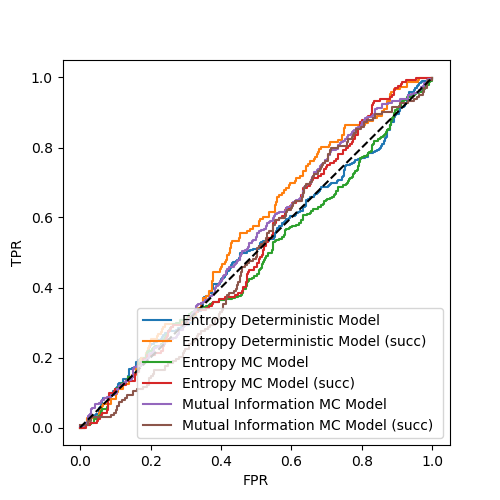}
  \caption{FGM with \(\epsilon=5\)}
  \label{fig:rcc_results_fgm_rn50roc_curves}
\end{minipage}
\begin{minipage}[t]{0.32\textwidth}
  \centering
  \includegraphics[width=\columnwidth]{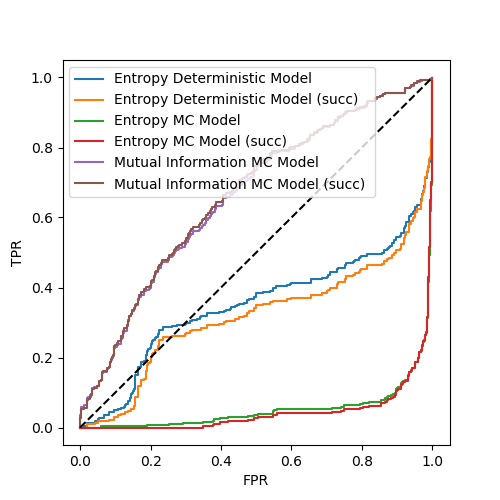}
  \caption{MIM with \(\epsilon=5\)}
  \label{fig:rcc_results_mim_rn50roc_curves}
\end{minipage}

\begin{minipage}[t]{0.32\textwidth}
  \centering
  \includegraphics[width=\columnwidth]{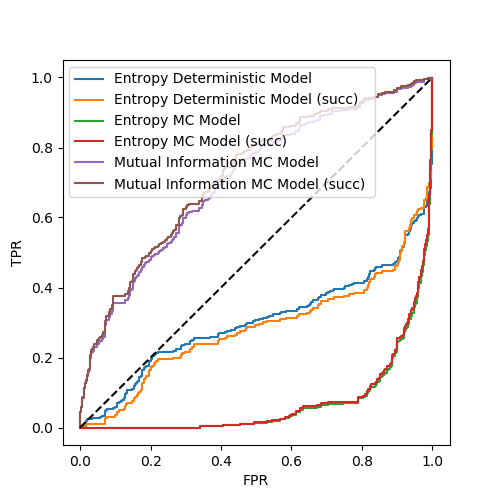}
  \caption{BIM with \(\epsilon=10\)}
  \label{fig:rcc_results_bim_rn50_1roc_curves}
\end{minipage}
\begin{minipage}[t]{0.32\textwidth}
  \centering
  \includegraphics[width=\columnwidth]{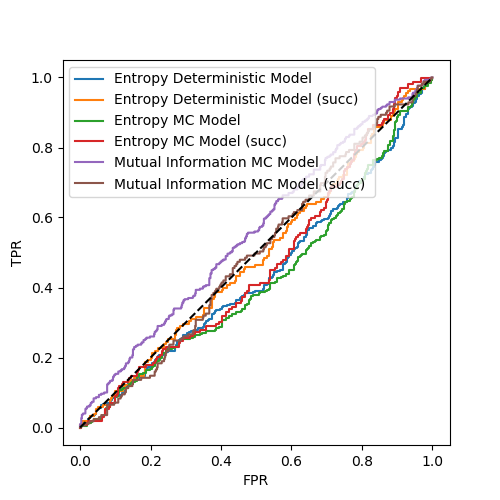}
  \caption{FGM with
    \(\epsilon=10\)}
  \label{fig:rcc_results_fgm_rn50_1roc_curves}
\end{minipage}
\begin{minipage}[t]{0.32\textwidth}
  \centering
  \includegraphics[width=\columnwidth]{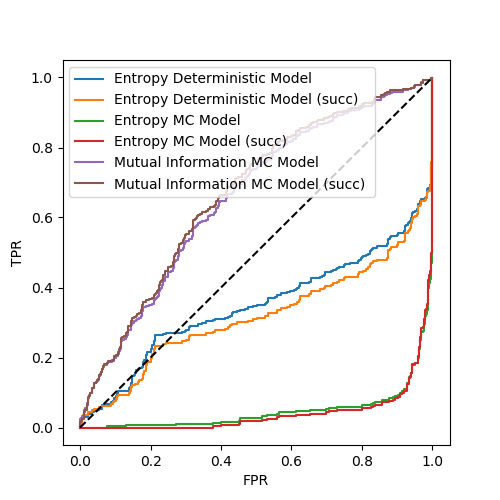}
  \caption{MIM with \(\epsilon=10\)}
  \label{fig:rcc_results_mim_rn50_1roc_curves}
\end{minipage}
\caption{ROC (precision recall) plots for adversarial example detection with different measures of uncertainty and different attacks. From left to right: basic iterative method (BIM), fast gradient method (FGM), and momentum iterative method (MIM). Top row uses $\epsilon$ of 5, bottom row uses $\epsilon$ of 10. All use infinity norm. (succ) denotes the quantity evaluated only for successful adversarial examples. We suspect that the low FGM attack success rate is related to the  difficultly we observe in identifying these using model uncertainty, however further investigation is required.}
\vspace{-4mm}
\end{figure*}

It has been observed by \citep{carlini2017adversarial} that many
proposed defences against adversarial examples fail to generalize from
MNIST. Therefore, we also evaluate the various uncertainty measures on a more
realistic dataset; the ASSIRA cats and dogs dataset (see Figure
\ref{fig:adv_examples} for example images).  The task is to distinguish
pictures of cats and dogs. While this is not a state of the art
problem, these are realistic, high resolution images.  We finetune a ResNet model  \citep{he2015deep}, pre-trained on
Imagenet, replacing the final layer with a dropout layer followed by a new fully connected layer. We use 20 forward passes for the Monte Carlo
dropout estimates.
We use dropout only on the layers we retrain,
treating the pre-trained convolutions as deterministic. 

We compare the receiver operating characteristic (ROC) of the predictive
entropy of the deterministic network, the predictive entropy of the
dropout network (equation \ref{predent}), and the MI of the dropout
network (the MI is always zero if the model is deterministic; this
corresponds to the approximating distribution \(q\) being a delta
function). Note that we compare with the same set of weights (trained
with dropout) --  the only difference is whether we use dropout at test
time. For each measure of uncertainty we generate the ROC plot by thresholding the uncertainty at different values, using the threshold to decide whether an input is adversarial or not.

The receiver operating characteristic is evaluated on a synthetic
dataset consisting of images drawn at random from the test set and
images from the test set corrupted by Gaussian noise, which comprise
the negative examples, as well as adversarial examples generated with the
Basic Iterative Method \citep{kurakin2016adversarial}, Fast Gradient
method \citep{goodfellow2014explaining}, and Momentum Iterative Method
\citep{dongboosting}. We test with the final attack because it is
notably strong, winning the recent NIPS adversarial attack
competition, and is simpler to adapt to stochastic models than the
other strong attacks in the literature, such as that of Carlini and
Wagner \citep{carlini2017towards}.

We find that only the mutual information gets a useful AUC on
adversarial examples.  In fact, 
most other measures of uncertainty seem to be worse than random guessing; this suggests that this dataset has a lot of examples
the model considers to be ambiguous (high aleatoric uncertainty), which mean
that the entropy has a high false positive rate. The fact the AUC of the entropy is low suggests that
the model is actually \textit{more} confident about adversarial examples than natural ones under this measure. 

An interesting quirk of this particular model is that the accuracy of using Monte
Carlo estimation is actually lower than the point estimates, even though the uncertainty estimates are sensible. Possibly this is because the dropout probability is quite high; only a subset of the features in the later layers of a convnet are
relevant to cat and dog discrimination, so this may be a relic of our transfer learning procedure; dropout does not normally have an adverse effect on the accuracy of fully trained models.

\section{Discussion \& Conclusion}

While security considerations for practical applications are clearly a
concern, fundamentally, we are interested in the question of whether
adversarial examples are an \textit{intrinsic} property of neural
networks, in the sense that they are somehow a property of the
function class, or whether they are an artefact of our training
procedure, that can be addressed through better training and
inference. We believe that the results in this paper are evidence in
favour of the latter; even approximate marginalization produces an
improvement in terms of robustness to adversarial examples.  It
is notable that fundamentally, these techniques can be derived without
any explicit reference to the adversarial setting, and no assumptions
are made about the distribution of adversarial examples. Rather,
looking for better uncertainty results in better robustness
automatically.

We do not claim that dropout alone provides a very convincing defence
against adversarial attack. Our results (in agreement with previous
literature on the subject) show that dropout networks are \textit{more
  difficult} to attack than their deterministic counterparts, but
attacks against them can still succeed while remaining imperceptible
to the human eye, at least in the white-box setting we
investigated. 

However, we think that this is likely to be because dropout is a fairly crude 
approximation that underestimates the uncertainty significantly, as suggested by our visualisations in 
latent space. These insights suggest that the best way to combat adversarial 
examples is from first principles; by improving model robustness and dealing with 
uncertainty properly, models become harder to fool as a side effect. Looking for 
scalable ways to improve on the uncertainty quality captured by dropout is an important avenue for future research in this area.

\FloatBarrier
\bibliography{paper} 
\bibliographystyle{apacite}

\end{document}